\documentclass[11pt]{article}

\usepackage[final]{acl}

\usepackage{times}
\usepackage{latexsym}
\usepackage{float}
\usepackage[T1]{fontenc}
\usepackage[utf8]{inputenc}
\usepackage{microtype}
\usepackage{inconsolata}
\usepackage{graphicx}
\usepackage{listings}
\usepackage{xcolor}

\usepackage{listings}
\usepackage{xcolor}

\usepackage{tabularx}
\usepackage{array}

\definecolor{jsonbg}{HTML}{FBFBFB}
\definecolor{jsonrule}{HTML}{D0D0D0}
\definecolor{jsonkey}{HTML}{0451A5}   
\definecolor{jsonstr}{HTML}{A31515}   
\definecolor{jsonnum}{HTML}{098658}   

\definecolor{jsonbg}{HTML}{F7F7F8}
\definecolor{jsonrule}{HTML}{D8D8DC}
\definecolor{jsonline}{HTML}{9A9AA2}
\definecolor{jsonkey}{HTML}{0451A5}    
\definecolor{jsonval}{HTML}{A31515}    
\definecolor{jsonpunct}{HTML}{444444}  

\lstdefinelanguage{json}{
  basicstyle       = \ttfamily\scriptsize\color{jsonpunct},
  string           = [s]{"}{"},
  stringstyle      = \color{jsonkey},
  comment          = [l]{:},
  commentstyle     = \color{jsonval},
  showstringspaces = false,
  columns          = fullflexible,
  keepspaces       = true,
  breaklines       = true,
  breakatwhitespace= true,
  breakindent      = 2em,
  postbreak        = \mbox{\textcolor{jsonline}{$\hookrightarrow$}\space},
  numbers          = left,
  numberstyle      = \ttfamily\tiny\color{jsonline},
  numbersep        = 9pt,
  xleftmargin      = 2.4em,
  framexleftmargin = 2.0em,
  frame            = single,
  rulecolor        = \color{jsonrule},
  backgroundcolor  = \color{jsonbg},
  framesep         = 6pt,
  aboveskip        = 1em,
  belowskip        = 1em,
}

\usepackage{longtable}
\usepackage{capt-of}
\usepackage{amssymb}
\usepackage{tabularx,makecell,multirow}
\usepackage{booktabs}
\usepackage{pifont}
\usepackage{enumitem}
\usepackage[most]{tcolorbox}
\usepackage{listings}
\usepackage{listingsutf8}
\usepackage{xspace}
\usepackage{cleveref}
\crefname{figure}{Figure}{Figures}
\Crefname{figure}{Figure}{Figures}
\crefname{table}{Table}{Tables}
\Crefname{table}{Table}{Tables}
\usepackage{stfloats}
\usepackage{mathtools}
\usepackage{colortbl}
\usepackage{xcolor}
\usepackage{tikz}
\usetikzlibrary{arrows.meta,calc,fit,positioning,shapes.geometric}

\newcommand{\dl}[1]{}
\newcommand{\aff}[1]{\textsuperscript{\textbf{#1}}}
\newcommand{\affmention}[1]{\textsuperscript{#1}}

\makeatletter
\providecommand{\@LN}[2]{}
\providecommand{\@LN@col}[1]{}
\makeatother

\newtcolorbox{promptbox}[1][Prompt]{
    colback=gray!5,
    colframe=orange!80!black,
    fonttitle=\bfseries,
    title=#1,
    rounded corners,
    boxrule=1.5pt,
    left=4mm,
    right=4mm,
    top=3mm,
    bottom=3mm,
    breakable  
}

\lstdefinestyle{promptcode}{
    basicstyle=\footnotesize\ttfamily,
    breaklines=true,
    backgroundcolor=\color{gray!3},
    frame=single,
    framesep=3pt,
    xleftmargin=10pt,
    xrightmargin=10pt
}

\title{Adaptive Influence Graphs for Failure Attribution \\ in Multi-Agent Systems}

\author{
   Yarden Bakish\aff{1}\thanks{Equal contribution.}\thanks{Work done during an
  internship at AWS Agentic AI.} \quad
  Amir Dudai\aff{2}\footnotemark[1]\thanks{Correspondence to:
  \texttt{dudaia@amazon.com}} \quad
  Roy Ganz\aff{2} \quad Oren Nuriel\aff{2} \\
  \textbf{Elad Ben Avraham}\aff{2} \quad \textbf{Mor Shpigel Nacson}\aff{2} \quad
  \textbf{Ron Litman}\aff{2} \\
  \affmention{1}Tel Aviv University \quad \affmention{2}AWS Agentic AI \\
}

\begin{document}
\maketitle

\begin{abstract}

Multi-agent LLM systems are increasingly deployed in real-world applications, where failures can be costly and difficult to localize. Despite growing efforts to automate failure attribution, diagnosing failed runs still largely relies on human engineers. Yet engineers rarely debug complex systems by reading raw logs end to end. Instead, observability tools organize traces around components, actions, and dependencies to support targeted navigation. We hypothesize that modern LLMs can benefit from the same paradigm. 
To test this hypothesis, we introduce \textbf{Adaptive Influence Graphs (AIGs)}, a two-stage agentic framework that first transforms a failed trace into a structured graph and then navigates it to identify the critical error. 
Across multiple models, we show that richer trace representations consistently improve failure attribution, with adaptive graph construction and agent-directed traversal yielding the strongest results.
AIGs establish a new state of the art on Who\&When, the standard benchmark for multi-agent failure attribution. This affirms our hypothesis that attribution depends not only on the diagnosing model, but also on how the trace is represented and explored.

\end{abstract}
\section{Introduction}
\label{sec:introduction}
\begin{figure}[t]
 \centering
 \includegraphics[width=\columnwidth]{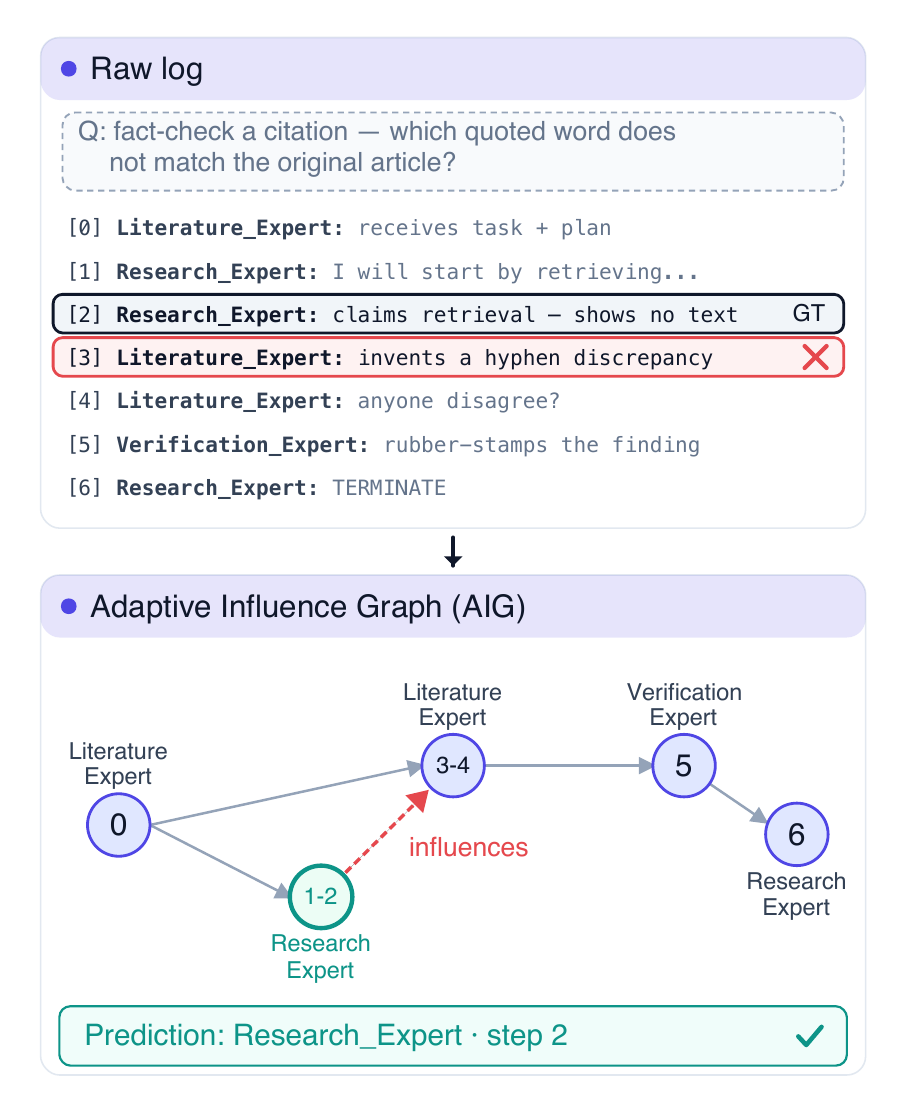}
 \caption{
   The same trace yields different attributions depending on its representation.
   The raw log highlights the visible symptom (step 3); the Adaptive Influence
   Graph traces the error back to its origin (step 2).
 }
 \label{fig:teaser}
\end{figure}

Multi-agent systems are increasingly deployed to solve complex real-world tasks, with executions spanning specialized agents, tools, and intermediate artifacts. When these systems fail, diagnosis still largely falls to human engineers, who must inspect long, interleaved logs to determine where the execution first went wrong. This process becomes increasingly costly as agentic systems grow more complex and operate over longer horizons, making automated diagnosis essential for scalable development and deployment. Who\&When formalized this problem as \emph{failure attribution}: identifying the failure-responsible agent and the decisive error step in a failed execution~\citep{whoandwhen2025,wang2026surveytrajectory}. We illustrate the task in the upper panel of \Cref{fig:teaser}.

Prior work on multi-agent failure attribution has primarily focused on improving the reasoning procedure applied to an execution trace, for example through multi-stage diagnosis pipelines~\citep[e.g.,][]{whoandwhen2025,raffles2025,a2p2025}. A complementary line of work restructures the trace before attribution using structured records or dependency representations~\citep{famas2025,cdcmas2025,li2026selfimprovingerrordiagnosismultiagent,rafi2026falattracingfailuresllm}. CHIEF~\citep{chief2026} further reconstructs flat logs as hierarchical causal graphs, organizing attribution through a predefined hierarchy and edge schema. Thus, existing approaches either reason over a largely fixed trace interface or restructure it using a predefined representation.

Human engineers rarely debug complex systems by reading raw logs end to end.
Instead, they rely on agent observability systems%
\footnote{Examples include LangSmith, Phoenix, and Amazon Bedrock AgentCore
Observability~\citep{langsmith,phoenix,agentcoreobservability}.}
that organize executions around components, actions, and dependencies,
enabling targeted traversal of the trace.
As modern agentic LLMs become increasingly capable of planning, tool use, and
iterative reasoning, we ask whether they can benefit from the same paradigm.
We hypothesize that failure attribution is partly an
\emph{interface-design} problem: it depends not only on the diagnosing model,
but also on how the trace is represented and traversed.
\Cref{fig:teaser} illustrates this hypothesis: on the same failed execution, the same reader misattributes the failure from the raw log but identifies the correct critical step from the structured graph.

\begin{figure}[t]
 \centering
 \includegraphics[width=\columnwidth]{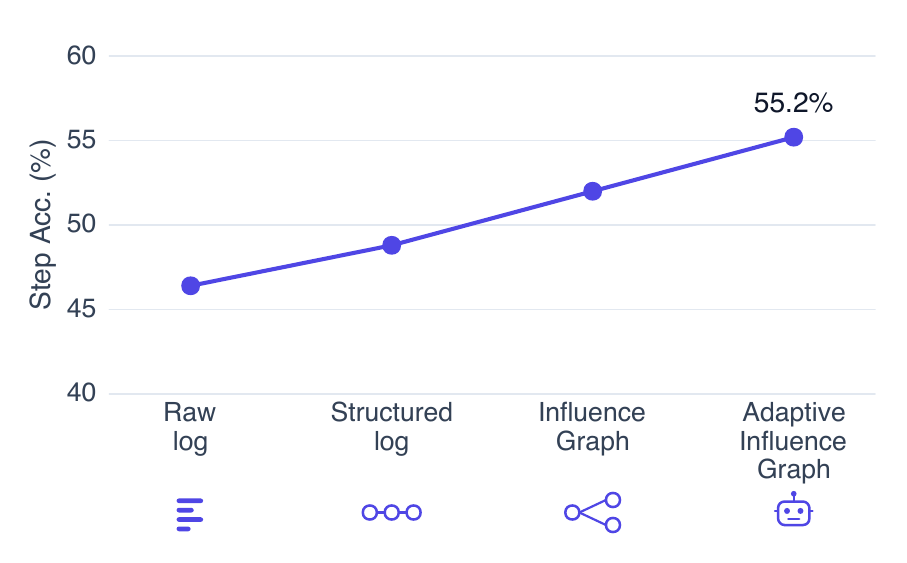}
\caption{
Step accuracy improves with richer trace representations, and improves further
when the adaptive graph is traversed by an agentic reader
(Algorithm-Generated Who\&When, Opus-5).
}
 \label{fig:ladder}
\end{figure}

To test this hypothesis, we separate failure attribution into two components:
how the trace is \emph{represented} and how that representation is \emph{read}.
We first evaluate a four-level representation ladder while holding the reader
fixed as a single call over the complete serialized interface
(\Cref{fig:ladder}):
(i) the raw log;
(ii) a structured log that groups consecutive steps from the same agent into
segments;
(iii) an Influence Graph that summarizes what each segment received and
produced and adds edges tracing how earlier outputs were inherited downstream;
and
(iv) an Adaptive Influence Graph, whose agentic builder chooses the action
boundaries and influence structure separately for each failed trace.
A critic--refiner audits the resulting adaptive graph for structural and
semantic consistency.
We then replace the single-call reader with an agentic reader. The graph
provides a high-level map of the execution, while the underlying log is exposed
only on demand. The reader iteratively chooses which steps to inspect, follows
influence links, and verifies the graph's claims before predicting the critical
agent and step. This design isolates the benefit of richer trace
representations from the additional benefit of agent-directed traversal.

We evaluate this design on both partitions of Who\&When, using the
Algorithm-Generated partition for controlled ablations. There, with Opus-5,
exact step accuracy rises from 46.40\% on raw logs to 48.80\% with structured
logs and 52.00\% with influence graphs; adaptive graph construction paired
with agent-directed traversal raises it further to 55.20\%
(\Cref{fig:ladder}). This establishes a new state of the art on the
Algorithm-Generated partition, surpassing the previous best reported result of
51.60\%, while also reaching 71.20\% agent accuracy and localizing 84.00\% of
failures within a $\pm 3$-step window. 
We further demonstrate the robustness of AIGs across models by evaluating them
with DS-V3.2, Sonnet-4, and GPT-5.6-sol in addition to Opus-5.
Our analysis shows that the gains concentrate on failures later in the trace,
where AIGs substantially improve exact step localization over raw-log reading.
Together, these findings provide
empirical support for our hypothesis that failure attribution depends not only
on the diagnosing model, but also on how the trace is represented and
traversed.

\paragraph{Contributions.}
\textbf{Failure attribution as interface design.}
We systematically demonstrate that multi-agent failure attribution depends not
only on the diagnosing model, but also on how the execution trace is
represented and traversed. A controlled representation ladder isolates the
benefit of progressively richer structure, while evaluations across multiple
models show that the effect is not specific to a single reasoner.
\textbf{Adaptive Influence Graphs.}
We introduce AIGs, a two-stage agentic framework that adaptively constructs an
influence graph and traverses it while verifying evidence against the original
log. The complete system establishes a new state of the art on the Algorithm-Generated Who\&When partition.

\section{Related Work}
\label{sec:related_work}

\paragraph{Failure attribution in multi-agent systems.}
A growing line of work studies how to diagnose failed executions by identifying the agent and step that first caused the error, where step-level accuracy remains limited~\citep{wang2026surveytrajectory}. The Who\&When benchmark~\citep{whoandwhen2025} formalizes the setting, and its prompted baselines differ only in how the log is sliced. Subsequent methods improve the \emph{inference procedure} applied to that transcript---scaffolding it with abductive prompts~\citep{a2p2025}, distributing judgment across a judge and criterion-specific evaluators~\citep{raffles2025}, compressing context by positional distance from a candidate step and voting across analyst personas~\citep{echo2025}, retrieving condensed error knowledge from previously annotated traces~\citep{correct2025}, verifying predefined error hypotheses against the full trajectory before localizing responsible agents~\citep{qiao2026verifymashypothesisverificationfailure}, or using model-based attribution through prefill-stage signals~\citep{liu2026masprismlightweightfailureattribution} and synthetic-failure fine-tuning~\citep{agentracer2025}. In each case the diagnosing model still receives the trace as a flat transcript, modified at most by step indexing, a fixed slicing policy, or distance-based truncation. Recent work broadens the task itself, releasing the inputs and tool logs each step saw~\citep{traceelephant2026}, allowing several defensible attributions per failure~\citep{mpbench2026}, introducing a cross-domain benchmark of first unrecoverable failures~\citep{barke2026agentrxdiagnosingaiagent}, or moving failure analysis online to detect decisive errors during execution~\citep{zhang2026agentforesightonlineauditingearly}. We work in Who\&When's partially observable setting but focus on a different axis: not only how a model reasons over a trace, but how that trace is represented before reasoning begins. Our method is also training-free: both the builder and the reader are prompted frontier models.

\paragraph{Restructuring the trace.}
A second line of work does restructure the trace before attribution. FAMAS~\citep{famas2025} abstracts each log record into a canonical $\langle$agent, action, state$\rangle$ triple and attributes via spectrum analysis; CDC-MAS~\citep{cdcmas2025} derives a data-dependency graph by a deterministic rule and recovers a causal DAG by constraint-based discovery; and StepFinder~\citep{stepfinder2026}, arguing that noisy logs interfere with LLM reasoning, encodes each step into a feature vector and trains a small sequential model, removing the LLM from attribution altogether. Closest to our work, prior methods construct structured dependency representations and search them for failure sources~\citep{chief2026,li2026selfimprovingerrordiagnosismultiagent,rafi2026falattracingfailuresllm}. These systems and related agent-debugging work establish that restructuring the trace helps~\citep{chen2026failedtrajectoriesreliablellm,li2026codetracertraceableagentstates,kang2026knowledgebasedzeroreplaydebuggingmultiagent}. Across these methods, however, the construction policy is fixed a priori through predefined record types, node boundaries, hierarchies, or relation vocabularies, whereas our builder jointly chooses node granularity and edge topology per trace. More broadly, LLM performance is sensitive to the position and spacing of relevant evidence, prompt formatting, and graph encoding~\citep{liu2023lostmiddlelanguagemodels,tian2025distancerelevantinformationpieces,sclar2024quantifyinglanguagemodelssensitivity,fatemi2023talklikegraphencoding}. 

\paragraph{Trace observability.}
Our approach is inspired by how humans debug agent systems in practice. Rather than reading a raw log top to bottom, they use observability platforms such as Arize Phoenix~\citep{phoenix}, built on OpenTelemetry~\citep{opentelemetry}, which ingest an execution trace and present it as a hierarchical, filterable object: spans nested by call structure, attributes to group and search on, and a tree to be expanded on demand. Adaptive Influence Graphs carry that design from a human consumer to an automated one: the builder acts as the instrumentation layer, turning the trace into a navigable graph, and the reader drills into its nodes and edges much as an engineer drills into spans.

\section{Method}
\label{sec:methods}

A trajectory is a sequence of $T$ steps
\[
(a_1, s_1), \ldots, (a_T, s_T),
\]
where $a_t$ is the agent acting at step $t$ and $s_t$ is the emitted log
content; the same agent may act at multiple steps. Given the original
task query $q$ and a failed final answer, the system must identify the
agent and step $(a^\star, t^\star)$ that first derailed the trajectory
from solving the task correctly. We evaluate exact agent accuracy and
exact step accuracy.

\begin{figure*}[t]
 \centering
 \includegraphics[width=\textwidth]{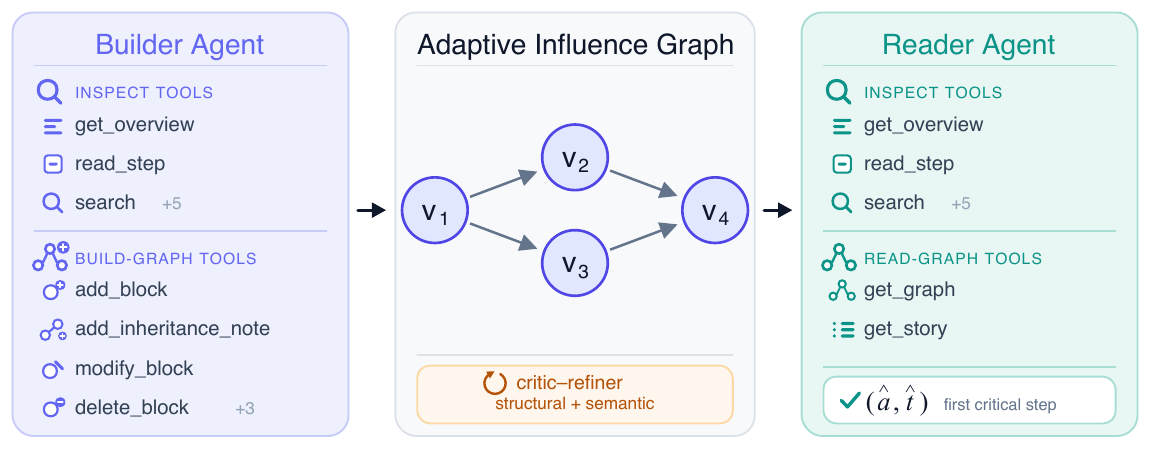}
 \caption{Build-read pipeline for Adaptive Influence Graphs (AIGs). The builder uses log-inspection and graph-construction tools to produce $G$; a critic--refiner audits its structural and semantic validity; and a reader uses graph and log-inspection tools to produce the final attribution.}
 \label{fig:methods_overview}
\end{figure*}

We decompose attribution into two stages: a \emph{builder} (\cref{subsec:build}) constructs a
structured representation $G$ from the query and trajectory, and a
\emph{reader} (\cref{subsec:read}) uses a set of tools to produce an attribution $(\hat{a}, \hat{t})$, as
shown in \cref{fig:methods_overview}. The stages are coupled: the reader
navigates $G$ and accesses the underlying log entries referenced by its
nodes on demand. Accordingly, $G$ is designed as an interface rather
than a summary, with inheritance edges for backtracking and nodes
grounded in the original log.

\subsection{Build: Constructing the Graph}
\label{subsec:build}

We evaluate builders from deterministic raw-log mapping to fully agentic graph construction.

\paragraph{Builder 1: Structured logs.}
The first structured builder is deterministic and LLM-free. It groups consecutive steps from the same agent into one node, yielding a chain of nodes $v_1, \dots, v_K$ in execution order, each node holding the verbatim steps of one agent's uninterrupted turn. This changes the surface form while preserving the evidence, so gains isolate the value of explicit structure rather than content.

\paragraph{Builder 2: Influence Graphs.}
The second builder uses an LLM to enrich this chain into an \emph{Influence Graph} (IG), adding node fields and inheritance edges in a single pass while leaving the node partition untouched. Each node receives a four-field abstraction $\alpha(v_k) = \langle \textit{summary},\ \textit{input},\ \textit{output},\ \textit{authorship} \rangle$---what the node did end-to-end, the input it received, the output it generated, and the node's role in producing that output---in addition to its raw content. The LLM also adds a directed edge from an earlier node to a later one when the later node reused part of the earlier node's work in a way tied to what went wrong. Each edge is annotated with two fields: \textit{inherited}, naming the element carried over, and \textit{effect}, describing how it went wrong downstream. An edge is drawn only under inheritance, so a node that starts fresh induces none; the edge set is sparse and may be empty because unsupported causal links can mislead the reader. See \cref{app:ig_example} for an example.

\paragraph{Builder 3: Adaptive Influence Graph.}
The Adaptive Influence Graph (AIG) generalizes the IG by replacing its fixed, single-pass construction procedure with an agentic builder. Both representations use step-grounded nodes and annotated inheritance edges. However, whereas the IG enriches a node partition fixed in advance, the AIG builder jointly determines the node boundaries and roles, fills their role-specific abstractions, and constructs the inheritance topology for each trace. The topology is therefore not fixed a priori and may connect non-adjacent actions, subject to the graph validity constraints described below.

The builder starts from an empty graph and interacts with the trace through two tool families. Inspection tools \(\mathcal{T}_{\mathrm{log}}\) expose selected parts of the trajectory without modifying the graph, while construction tools \(\mathcal{T}_{\mathrm{build}}\) create, modify, and delete nodes and inheritance edges. The builder interleaves inspection and construction until it submits the completed graph. App.~\ref{app:tools} lists the full tool set.

The builder outputs a directed graph \(G=(V,E)\), where \(V\) is the set of nodes and \(E\) is the set of directed edges. Every node is grounded in one or more raw-log steps and assigned a type from a fixed vocabulary. The node type determines which fields the builder must provide.

In the single-type variant, every node uses the same four fields as the IG---\textit{summary}, \textit{input}, \textit{output}, and \textit{authorship}---and inheritance edges use the same annotations. This preserves the IG schema while allowing the agentic builder to choose the node boundaries and graph topology.

The agentic framework benefits further from enlarging this output space to typed, grounded graphs. We extend the fixed node vocabulary with additional roles, each governing a distinct node type; per-role descriptions appear in App.~\ref{app:typed_nodes}.

\textbf{Critic--Refiner Loop.}
Admissible graphs must satisfy a validity predicate $\Phi(G)$, including connectivity, acyclicity, one root, and one sink. After submission, a critic audits the graph for at most $R$ rounds, checking structural
violations programmatically against $\Phi$ and semantic violations against the log, such as a node that is unfaithful to its referenced steps. A refiner applies minimal edits to resolve any violations, and the loop stops once none remain or the round budget is exhausted. See \cref{app:aig_example} for an example.

\subsection{Read: Attributing Failure from the Representation}
\label{subsec:read}

We evaluate two readers that differ in how they consume the representation.

\paragraph{Single-call reading.}
The single-call reader receives the serialized representation along with the raw log and emits one attribution.

\paragraph{Agentic graph reading.}
The agentic reader retrieves the finalized graph \(G\), then selectively inspects the underlying trajectory \(\tau\) using its inspection tools \(\mathcal{T}_{\mathrm{log}}\). Starting from a candidate node from the graph, it works backward over incoming inheritance edges. We instruct it to move blame from the current node to an edge's source only when the raw log confirms that the current node's output exhibits the downstream error described by the edge's \textit{effect} annotation. This process continues until no incoming inheritance edge explains the observed failure. The reader then resolves the selected node to an exact agent and step using its step references and the raw log, and submits \((\hat{a},\hat{t})\).

\subsection{Implementation Details}
\label{subsec:implementation}

Each stage is a separate agent built with the Strands Agents SDK v1.37~\citep{strands}, where tools are Python functions whose schemas the SDK derives and drives, and the graph is shared mutable state that construction tools edit in place and \texttt{read\_graph}() serializes on demand. We run the full AIG pipeline with the following backbones---Opus-5, Sonnet-4, and DeepSeek-V3.2 (DS-V3.2)---and repeat the representation ablations under each, holding the model fixed within a ladder. Reported results are Opus-5 unless stated otherwise. Models are accessed via the Amazon Bedrock Converse API (Sonnet-4 and DS-V3.2 at temperature $0.6$; Opus-5 and GPT-5.6-sol at their defaults, which cannot be set), with no task-specific fine-tuning. The critic--refiner loop runs at most $R=3$ rounds, exiting early on an empty violation set. A stage ends at \texttt{submit}.

\section{Results}
\label{sec:results}

We evaluate Adaptive Influence Graphs on both partitions of
Who\&When~\citep{whoandwhen2025}. We first report benchmark performance and
consistency across models, then use a controlled interface ladder on the
Algorithm-Generated partition to isolate the contributions of trace
representation and agent-directed traversal. \Cref{sec:analysis} examines
where the gains occur and how the builder and reader use the resulting
interface.

\subsection{Who\&When Benchmark Results}
\label{subsec:benchmark_results}

\begin{table*}[!t]
\centering
\small
\setlength{\aboverulesep}{0pt}
\setlength{\belowrulesep}{0pt}
\renewcommand{\arraystretch}{1.25}
\definecolor{oursgray}{gray}{0.92}
\begin{tabularx}{\textwidth}{>{\raggedright\arraybackslash}X>{\raggedright\arraybackslash}p{2.0cm}cccc}
\toprule
\multirow{2}{*}{\textbf{Method}} & \multirow{2}{*}{\textbf{Model}} & \multicolumn{2}{c}{\textbf{Algorithm-Generated Dataset}} & \multicolumn{2}{c}{\textbf{Hand-Crafted Dataset}} \\
\cmidrule(lr){3-4} \cmidrule(lr){5-6}
 & & \textbf{Step Acc. $\uparrow$} & \textbf{Agent Acc. $\uparrow$} & \textbf{Step Acc. $\uparrow$} & \textbf{Agent Acc. $\uparrow$} \\
\midrule
All-at-Once & Opus-5 & 46.40 & 67.20 & 22.41 & 50.00 \\
Step-by-Step & Opus-5 & 42.40 & 54.40 & 25.86 & 58.62 \\
\midrule
CDC-MAS~\citep{cdcmas2025} & GPT-4o & 36.00 & 48.50 & 18.20 & 56.80 \\
AgenTracer-8B~\citep{agentracer2025} & Qwen3-8B & 37.30 & 63.73 & 20.68 & 63.82 \\
CORRECT~\citep{correct2025} & GPT-5 & 38.10 & -- & 17.20 & -- \\
CHIEF$^\dagger$~\citep{chief2026} & Opus-5 & 45.24 & 64.29 & 18.97 & 51.72 \\
CHIEF~\citep{chief2026} & DS-V3.2 & 45.60 & 68.80 & \textbf{29.31} & \textbf{72.41} \\
A2P~\citep{a2p2025} & GPT-4o & 47.50 & 65.40 & \textbf{29.31} & 58.62 \\
RAFFLES~\citep{raffles2025} & Llama-3.3-70B & 43.65 & -- & 20.69 & -- \\
RAFFLES~\citep{raffles2025} & Sonnet-4 & 51.60 & -- & 22.41 & -- \\
\midrule
\rowcolor{oursgray} Adaptive Influence Graph & DS-V3.2 & 53.17 & 65.08 & 24.14 & 60.34 \\

\rowcolor{oursgray} Adaptive Influence Graph & Sonnet-4 & 53.97 & 67.46 & \textbf{29.31} & 53.45 \\
\rowcolor{oursgray} Adaptive Influence Graph & GPT-5.6-sol & 	54.76  & 67.46 & 24.14 & 55.17	 \\

\rowcolor{oursgray} Adaptive Influence Graph & Opus-5 & \textbf{55.20} & \textbf{71.20} & \textbf{29.31} & 63.79 \\

\bottomrule
\end{tabularx}
\caption{Who\&When benchmark. Rows use no ground-truth access at inference time. One Algorithm-Generated log is content-filtered by Opus-5; therefore, for this model we use $n{=}125$. $^\dagger$Run using the authors' released code. Our methods are shaded.}
\label{tab:sota_alg_generated_no_gt}
\end{table*}

\paragraph{Algorithm-Generated partition.}
AIGs with Opus-5 achieve 55.20\% step accuracy and 71.20\%
agent accuracy without the ground-truth answer at inference time.
This establishes a new state of the art on the Algorithm-Generated partition,
improving over the previous best step accuracy of 51.60\% by 3.6 percentage
points (\Cref{tab:sota_alg_generated_no_gt}).

\paragraph{Consistency across models.}
The result is not tied to Opus-5. AIGs reach 54.76\% step accuracy with
GPT-5.6-sol, 53.97\% with Sonnet-4, and 53.17\% with DS-V3.2; all three
surpass the previous best result. The advantage holds in matched-model
comparisons: AIGs outperform CHIEF by 9.96 points with Opus-5 and 7.57 points
with DS-V3.2, and RAFFLES by 2.37 points with Sonnet-4. It therefore
persists across model families and matched-model comparisons.

\paragraph{Hand-Crafted partition.}
With Opus-5, AIGs reach 29.31\% step accuracy on the Hand-Crafted partition,
matching A2P and CHIEF for the highest result without ground-truth access.
Their 63.79\% agent accuracy remains below CHIEF's 72.41\%, showing that the
step-level advantage does not extend to agent accuracy on this partition.

\subsection{The Interface Ladder}
\label{subsec:representation_ablations}

\begin{table}[t]
\centering
\small
\setlength{\aboverulesep}{0pt}
\setlength{\belowrulesep}{0pt}
\renewcommand{\arraystretch}{1.25}
\setlength{\tabcolsep}{3pt}
\begin{tabularx}{\linewidth}{>{\raggedright\arraybackslash}Xcc}
\toprule
\textbf{Representation} & \textbf{Step $\uparrow$} & \textbf{Agent $\uparrow$} \\
\midrule
Raw log & 46.40 & 67.20 \\
Structured log & 48.80 & 63.20 \\
Influence Graph & 52.00 & 70.40 \\
Adaptive Influence Graph$^{\text{LLM reader}}$ & 50.40 & 64.80 \\
\textbf{Adaptive Influence Graph} & \textbf{55.20} & \textbf{71.20} \\
\bottomrule
\end{tabularx}
\caption{Opus-5 representation ablations on the Algorithm-Generated Who\&When partition (no ground-truth access). Our AIGs are in bold.}
\label{tab:opus_ablations}
\end{table}

We next isolate how much of the improvement comes from the trace
representation and how much requires agent-directed traversal. On the
Algorithm-Generated partition, we hold Opus-5 fixed and progressively compare
the raw log, a structured log, an Influence Graph, an adaptively constructed
graph read as a single prompt, and the complete AIG system
(\Cref{tab:opus_ablations}).

\paragraph{Explicit structure improves attribution.}
Grouping consecutive steps from the same agent into segments raises exact step
accuracy from 46.40\% to 48.80\%, despite preserving the original evidence and
adding no semantic abstraction. This isolates the benefit of making the
execution structure explicit. The effect is even larger with Sonnet-4 and
DS-V3.2, whose accuracy rises from 20.63\% to 48.41\% and from 19.84\% to
45.24\%, respectively (App.~\ref{app:cross_model_ablations}). Once the trace is
structured, the three models perform within four points of one another, suggesting
that Opus-5 already recovers much of this structure from the raw log.

\paragraph{Influence structure adds further gains.}
Enriching the structured log into an Influence Graph raises exact step accuracy
from 48.80\% to 52.00\%. The representation augments each segment with a
compact account of what it received and produced, then connects segments with
inheritance edges that record how earlier outputs propagate downstream. The
reader still consumes the entire representation in a single call, so this gain
comes from making the execution's influence structure explicit rather than
from agent-directed traversal.

\paragraph{Agent-directed traversal unlocks the adaptive graph.}
When the adaptively constructed graph is flattened into a single prompt, step
accuracy reaches 50.40\%. Giving the reader control over how to traverse the
same graph raises accuracy by 4.8 points to 55.20\%, the strongest result in
the ladder. The graph therefore serves not merely as a richer prompt, but as a
high-level map that directs the reader toward relevant evidence and supports
selective verification against the original log. Overall, adaptive construction
paired with agent-directed traversal improves exact step accuracy by 8.8
points over raw-log reading. This interaction supports our central claim that
trace representation and reading should be designed jointly.

\section{Analysis}
\label{sec:analysis}

We next examine where AIGs improve attribution and how the two interface stages
produce that gain. We analyze which failures benefit most, then characterize
the builder's graphs and the reader's evidence gathering.

\subsection{Where AIGs Help}
\label{subsec:error_depth}

\paragraph{AIGs improve localization deeper in the trace.}
A paired Sonnet-4 analysis shows that the gains concentrate on later failures,
where downstream consequences can obscure their origin. Exact accuracy on
failures at step 5 or later rises from 10.8\% with raw-log reading to 40.5\%
with AIGs. Among incorrect raw-log predictions, 76\% still identify the correct
agent, suggesting that the main difficulty is precise step localization rather
than agent identification. This is consistent with AIGs helping the reader
trace downstream consequences to an earlier source. Opus-5 shows the same
qualitative pattern, with a smaller margin because its raw-log baseline is
already stronger.

\begin{table}[t]
\centering
\small
\setlength{\aboverulesep}{0pt}
\setlength{\belowrulesep}{0pt}
\renewcommand{\arraystretch}{1.25}
\resizebox{\columnwidth}{!}{%
\begin{tabular}{lcccc}
\toprule
\textbf{Representation} & \textbf{Step Acc.} & \textbf{$\pm$1} & \textbf{$\pm$2} & \textbf{$\pm$3} \\
\midrule
RAFFLES (Llama-3.3-70B) & 43.65 & 58.73 & 73.81 & 82.54 \\
\midrule
Structured log & 48.80 & 60.00 & 72.00 & 80.00 \\
Influence Graph & 52.00 & 61.60 & 73.60 & 80.00 \\
Adaptive Influence Graph & \textbf{55.20} & \textbf{64.00} & \textbf{78.40} & \textbf{84.00} \\
\bottomrule
\end{tabular}%
}
\caption{Step accuracy under relaxed tolerance windows ($\pm n$ steps) on the Algorithm-Generated Who\&When partition (Opus-5). RAFFLES results as published by the authors, who report tolerance windows only for Llama-3.3-70B.}
\label{tab:step_plus_n}
\end{table}

\paragraph{AIGs remain strongest under relaxed localization.}
Among methods reporting tolerance windows, AIGs lead at every window
(\Cref{tab:step_plus_n}): 55.20\% exactly, 64.00\% within one step, 78.40\%
within two, and 84.00\% within three. Even when missing the exact annotation,
they often localize the failure nearby.

\subsection{What the Builder Constructs}
\label{subsec:graph_properties}

\begin{figure}[t]
\centering
\includegraphics[width=\linewidth]{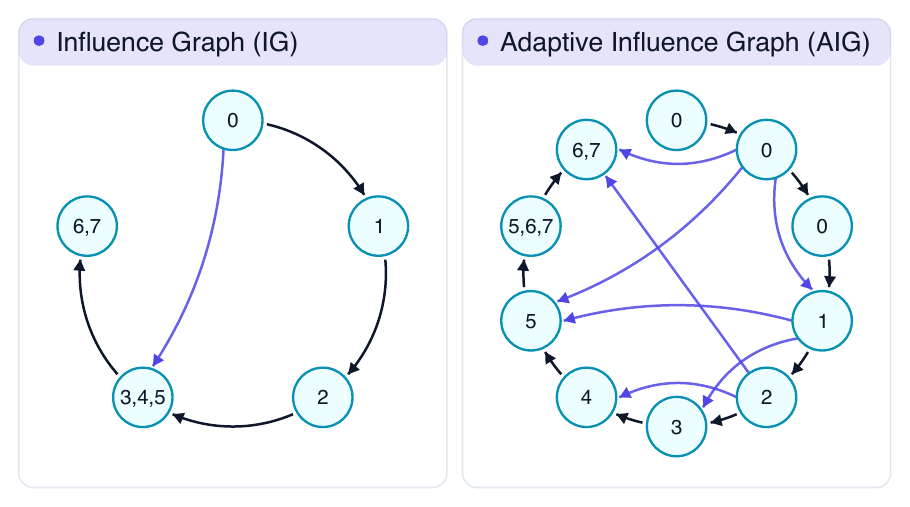}
\caption{The same trace encoded as an Influence Graph and as an Adaptive Influence Graph.}
\label{fig:graph_compare}
\end{figure}

Because AIGs are model-authored, we can examine how adaptive construction
changes the resulting representation. \Cref{tab:graph_properties} compares
AIGs with the fixed Influence Graph builder on the same failed traces, while
\Cref{fig:graph_compare} illustrates both representations for a single
execution. Edge counts include the sequential links between adjacent nodes and
the inheritance links added on top of them.

\begin{table}[t]
\centering
\small
\setlength{\tabcolsep}{4pt}
\begin{tabular*}{\linewidth}{@{\extracolsep{\fill}}l r@{\extracolsep{0pt}\,$\pm$\,}l@{\extracolsep{\fill}}r@{\extracolsep{0pt}\,$\pm$\,}l@{}}
\toprule
\textbf{Property} & \multicolumn{2}{c}{\textbf{AIGs}} & \multicolumn{2}{c}{\textbf{IGs}} \\
\midrule
Nodes per graph                    & 9.5  & 2.0  & 7.5  & 2.2  \\
Edges per graph                    & 13.5 & 3.6  & 9.0  & 3.2  \\
Edge density (edges / node)        & 1.4  & 0.2  & 1.2  & 0.2  \\
Max hub out-degree                 & 3.6  & 1.1  & 1.8  & 0.6  \\
Max steps merged into one node     & 2.8  & 1.2  & 1.6  & 0.5  \\
Mean cross-range distance          & 3.6  & 1.0  & 3.1  & 1.0  \\
Longest single cross-edge          & 5.6  & 2.1  & 4.2  & 2.2  \\
\bottomrule
\end{tabular*}
\caption{Properties of the constructed graphs averaged across the Algorithm-Generated partition ($\pm$ standard deviation across graphs).}
\label{tab:graph_properties}
\end{table}

\paragraph{Adaptive construction produces richer, variable-granularity graphs.}
AIGs contain more nodes and edges than fixed Influence Graphs on average
(9.5 vs. 7.5 nodes and 13.5 vs. 9.0 edges), yielding a denser
representation of the execution. Their most connected node also has twice the
out-degree (3.6 vs. 1.8), indicating that the builder often identifies one
upstream action whose output influences several later steps. At the same time,
AIGs merge longer runs of consecutive same-agent steps into single
action-level nodes: the most compacted node covers 2.8 steps on average,
compared with 1.6 for the fixed builder. Thus, the adaptive builder does not
apply uniformly finer or coarser segmentation; it varies granularity across
the trace while exposing broader patterns of downstream influence.

\paragraph{Adaptive graphs expose longer-range influence.}
Inheritance links in both span multiple steps, but AIGs reach farther: their
mean cross-range distance is 3.6 steps versus 3.1 for fixed Influence Graphs,
and their longest edge spans 5.6 steps on average versus 4.2. Together with the
higher hub out-degree, this suggests the adaptive builder connects an upstream
action to multiple downstream consequences rather than merely restating the
local execution order. The resulting graph explicitly encodes long-range
propagation for the reader.

\subsection{How the Reader Uses the Graph}
\label{subsec:reader_behavior}

\begin{figure*}[t]
\centering
\includegraphics[width=\textwidth]{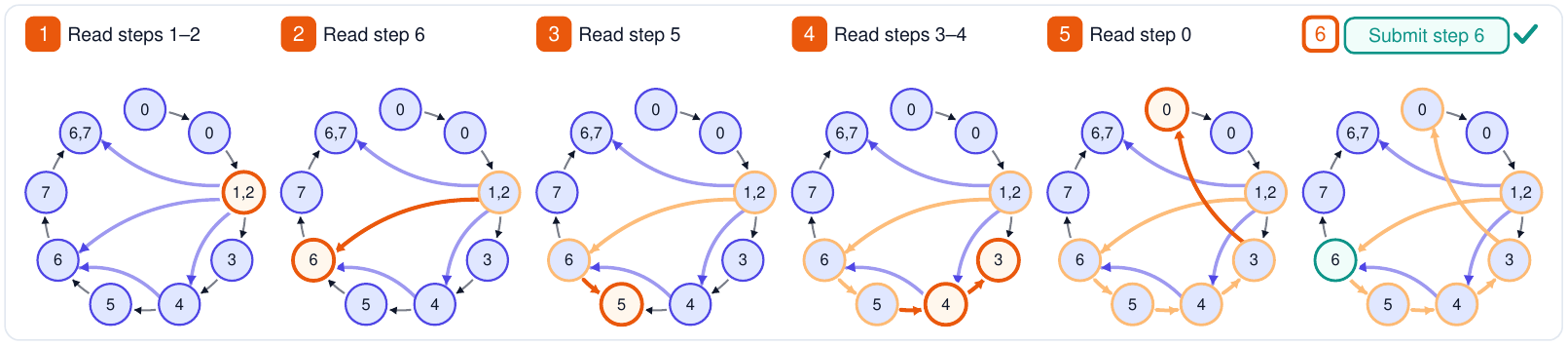}
\caption{Each panel highlights one move in orange; chords are inheritance edges and numbers are the raw-log steps a node merges. Rather than follow transcript order, the reader walks inheritance edges from the node the graph implicates, then commits to step 6, matching the annotation.}
\vspace{6pt}
\label{fig:reader_trajectory}
\end{figure*}

The reader's interaction traces let us examine how it uses the graph rather
than only whether the final prediction is correct.
\Cref{tab:reader_behavior} summarizes the reader executions behind the
Sonnet-4 results, and \Cref{fig:reader_trajectory} visualizes one complete
trajectory.

\begin{table}[t]
\centering
\small
\setlength{\tabcolsep}{4pt}
\begin{tabular*}{\linewidth}{@{\extracolsep{\fill}}lc@{}}
\toprule
\multicolumn{2}{@{}l}{\emph{Agentic reading profile}} \\
\midrule
First raw read is already $\hat{t}$              & 61.3\% \\
Opened steps within $\pm 1$ of $\hat{t}$         & 48.4\% \\
\midrule
Multi-step investigation                          & 86.5\% \\
Single-step verification                         & 11.9\% \\
Commits from the graph alone                     & 1.6\% \\
\bottomrule
\end{tabular*}
\caption{Reader behavior on the Algorithm-Generated partition (Sonnet-4, no
ground-truth access). The last three rows partition executions by reading depth.
$\hat{t}$ is the step the execution submits.}
\label{tab:reader_behavior}
\end{table}

\paragraph{The graph focuses, but does not replace, evidence gathering.}
The graph quickly directs the reader toward a candidate region: in 61.3\% of
executions, the first raw step inspected is also the submitted step, and
48.4\% of inspected steps lie within one step of the final prediction.
Nevertheless, the reader rarely accepts the graph without verification.
It investigates multiple raw steps in 86.5\% of executions and commits from
the graph alone in only 1.6\%. Thus, the graph acts as a high-level map that
focuses the search, while the reader chooses which evidence to inspect before
attributing the failure. \Cref{fig:reader_trajectory} illustrates this behavior:
the reader moves across influence-linked nodes, checks their underlying steps,
and commits to the annotated critical step.

Taken together, these analyses show that the two stages play complementary
roles: the builder exposes action-level and long-range influence structure,
while the reader uses that structure to focus a selective verification process
over the original log.

\section{Conclusion}
\label{sec:conclusion}

Failure attribution in multi-agent systems is not only about reasoning over
execution traces, but also representing them. We adapt the observability
practice of rendering a trace as a structured, navigable object
\citep{phoenix} to automated attribution: an agentic builder turns a failed
trace into an Adaptive Influence Graph, and an independent reader navigates it
while checking its claims against the underlying log.

Holding the model backbone fixed, enriching how the trace is represented and read moves step accuracy from 46.40\% to 55.20\% on the Algorithm-Generated Who\&When partition without ground-truth access. This is a new state of the art with +3.6 points over the best reported result~\citep{raffles2025}. AIGs retain their advantage across the backbones we test.

The interface ladder shows that explicit structure carries much of the
gain. Structured logs outperform raw transcripts, and influence graphs improve
further. The strongest result comes from pairing adaptive graph construction
with dynamic navigation. Analysis of the constructed graphs shows how this
adaptivity materializes. The builder merges consecutive steps into coarser
actions and draws inheritance edges between nodes several steps apart. The
reader follows these edges selectively and verifies graph claims against the
original log.

As agent traces grow longer and more interleaved, the interface to the trace
becomes a central part of the attribution method. These results show that
failure attribution depends not only on the diagnosing model, but also on the
interface through which it understands and explores the trace.

\section*{Limitations}

We evaluate on Who\&When, the standard benchmark for this task, across both of its partitions; whether the representation ladder transfers to other multi-agent frameworks and task distributions remains open. Who\&When exposes agent outputs but not the inputs each agent saw; on benchmarks that also release inputs and tool logs~\citep{traceelephant2026}, a builder could ground nodes in what each agent actually observed, which we expect to help rather than hinder. Following the benchmark protocol, we score the final attribution against a single annotated step, so we do not measure the intermediate graph directly: absent gold annotations for node granularity or inheritance edges, the builder's structure is validated by the critic and by downstream accuracy rather than against a reference graph. Finally, AIGs add a build stage and critic--refiner loop, costing more tokens than raw-log reading; the method is therefore best suited to settings where a reliable diagnosis is worth that overhead.

\section*{Ethical considerations}
This work studies failure attribution over execution traces from publicly available data; it involves no human subjects, no personal data, and no user-generated content beyond those benchmarks. The intended use is diagnostic---helping developers localize the agent and step responsible for a failed multi-agent run so they can debug it. As with any attribution method, outputs are imperfect and should not be treated as definitive assignments of blame, particularly in settings where an incorrect attribution could have major downstream consequences.


\bibliography{custom}

\clearpage
\appendix
\section{Tools}
\label{app:tools}

\Cref{tab:tools} lists the full tool surface available to the agentic
builder, $\mathcal{T}_{\mathrm{log}} \cup \mathcal{T}_{\mathrm{build}}$
(\cref{subsec:build}). The split reflects the two things a builder
does. Inspection tools query the trajectory $\tau$ at a chosen granularity---an
overview, one step, a range, a regex match---and leave the graph untouched;
the builder never receives $\tau$ in full unless it asks for it. Construction
tools are the only way the graph changes: they create and edit typed nodes,
ground them in log steps, and attach the $\langle\textit{inherited},
\textit{effect}\rangle$ annotations that make an edge an inheritance claim
rather than an adjacency. The reader is given a read-only subset: the graph accessors plus
$\mathcal{T}_{\mathrm{log}}$, so it can verify any node against the steps that
node cites, but cannot revise what it is reading.  \\

\begin{table*}[t]
\centering
\footnotesize
\setlength{\tabcolsep}{6pt}
\renewcommand{\arraystretch}{1.05}

\begin{tabularx}{\textwidth}{
  @{}
  >{\raggedright\arraybackslash}p{0.27\textwidth}
  >{\raggedright\arraybackslash}X
  @{}
}
\toprule
\textbf{Tool} & \textbf{Description} \\
\midrule

\multicolumn{2}{@{}l}{\emph{Inspection} ($\mathcal{T}_{\mathrm{log}}$)} \\
\midrule
\path{get_overview}
  & Compact metadata about the log: step count, per-step agent, and sizes. \\
\path{read_full_log}
  & Return every step concatenated. \\
\path{read_step}
  & Return one step's content. \\
\path{read_step_range}
  & Return an inclusive range of steps. \\
\path{list_agents}
  & List every agent and its system prompt. \\
\path{get_question}
  & Return the original user question verbatim. \\
\path{search}
  & Regex-search all step contents, returning up to 20 matches. \\
\path{list_block_kinds}
  & Return the closed block-kind vocabulary and required fields. \\
\path{get_story}
  & Render the current graph as Markdown. \\
\path{get_graph}
  & Return the current graph as JSON. \\

\midrule
\multicolumn{2}{@{}l}{\emph{Construction} ($\mathcal{T}_{\mathrm{build}}$)} \\
\midrule
\path{add_block}
  & Create a block (task, orchestrator, agent, or conclusion) from its fields
    and \texttt{from}/\texttt{to} lists; inheritance notes may accompany it. \\
\path{add_inheritance_note}
  & Attach a $\langle\text{bias},\text{anomaly}\rangle$ note to a direct
    source$\rightarrow$target edge. \\
\path{add_skip_attribution}
  & Attach a $\langle\text{bias},\text{anomaly}\rangle$ note to a non-adjacent
    source$\rightarrow$target propagation. \\
\path{modify_block}
  & Overwrite an existing block's fields; kind and ID are immutable. \\
\path{delete_block}
  & Delete a block and all edges, including notes, touching it. \\
\path{delete_inheritance_note}
  & Clear the inheritance notes on a direct edge. \\
\path{delete_skip_attribution}
  & Remove a non-adjacent skip attribution. \\
\path{mark_defect_recorded}
  & Flag the node whose \texttt{output} fails its optional system-produced
    \texttt{instruction\_or\_question} yet is committed downstream; record
    what fell short as the critical-step anchor. \\
\path{submit}
  & Emit the final prediction: agent, step, reason, and critical block.
    Called once. \\
\bottomrule
\end{tabularx}

\caption{\textbf{Builder tools.} Inspection reads the log $\tau$;
construction edits the graph.}
\label{tab:tools}
\end{table*}

\section{Typed Nodes}
\label{app:typed_nodes}

\Cref{tab:node_types} gives the node vocabulary $\Theta_V$
(\cref{subsec:build}) and the fields each type carries. The type is a
workflow role: \texttt{task} and \texttt{conclusion} are
the graph's two endpoints, fixed by the validity predicate $\Phi$ to one root
and one sink, while \texttt{orchestrator} and \texttt{agent} distinguish a turn
that routes work from a turn that performs it. Fields follow from the role. Only
\texttt{agent} nodes carry \textit{input} and \textit{output}, because only a
working turn has an instruction it received and a result it committed---and that
pair is what an inheritance edge connects, since what a later node inherits is
some earlier node's output. Every node carries $\gamma(v)$ as
\text{step\_refs}, so any field can be checked against the steps it claims to
abstract; this is what makes the reader's verification against $\tau$
(\cref{subsec:read}) possible at all. \\

\begin{table*}[t]
\centering
\footnotesize
\begin{tabularx}{\textwidth}{@{}l>{\raggedright\arraybackslash}X@{}}
\toprule
\textbf{Node type} & \textbf{Role \& fields} \\
\midrule
\texttt{task} &
The user's question that seeds the trajectory---the root of the graph
(no incoming edges). \newline
\textit{Fields:} \texttt{content} (the question), \texttt{operative\_brief}
(the instructions the trajectory starts under), \texttt{step\_refs} \\
\midrule
\texttt{orchestrator} &
A coordinating turn---an agent that plans, delegates, or judges progress
and routes the work onward. \newline
\textit{Fields:} \texttt{agent\_name}, \texttt{summary} (what the agent did
in this turn, start to end), \texttt{authorship} (as in \texttt{agent}),
\texttt{step\_refs} \\
\midrule
\texttt{agent} &
A working turn---an invoked agent that executes a delegated instruction
and produces a result. \newline
\textit{Fields:} \texttt{agent\_name}, \texttt{summary}, \texttt{input}
(the instruction, context, or prior result it worked from), \texttt{output}
(the concrete result, decision, tool output, or handoff it produced),
\texttt{authorship} (whether it authored that output),
\texttt{defect\_recorded} (the node whose \texttt{output} fails the optional
system-emitted \texttt{instruction\_or\_question} it was given, yet is
committed downstream), \texttt{step\_refs} \\
\midrule
\texttt{conclusion} &
The final answer the system produced---the terminal sink
(no outgoing edges). \newline
\textit{Fields:} \texttt{content} (the final answer), \texttt{step\_refs} \\
\bottomrule
\end{tabularx}
\caption{Node types in the agentic graph. Every node carries
\texttt{step\_refs}, the raw-log steps it covers.}
\label{tab:node_types}
\end{table*}

\section{With-Ground-Truth Results}
\label{app:with_gt_results}

\Cref{tab:sota_alg_generated_w_gt} reports the with-ground-truth setting, where
the attributor has access to the correct final answer during diagnosis.
Ground-truth access does not consistently help: AIG drops are at most four
traces, within sampling noise at these sample sizes. Similar regressions appear
wherever both settings are reported---in the benchmark's own baselines
\citep{whoandwhen2025}, in CHIEF's re-evaluation of them \citep{chief2026}, and
across the baselines of \citet{agentracer2025}, who note that ground truth
``may sometimes mislead the attribution process.''

\begin{table*}[t]
\centering
\small
\setlength{\tabcolsep}{4pt}
\setlength{\aboverulesep}{0pt}
\setlength{\belowrulesep}{0pt}
\renewcommand{\arraystretch}{1.25}
\definecolor{oursgray}{gray}{0.92}

\begin{tabularx}{\textwidth}{
  @{}
  >{\raggedright\arraybackslash}X
  >{\raggedright\arraybackslash}p{2.0cm}
  *{4}{>{\centering\arraybackslash}X}
  @{}
}
\toprule
\multirow{2}{*}{\textbf{Method}} &
\multirow{2}{*}{\textbf{Model}} &
\multicolumn{2}{c}{\textbf{Algorithm-Generated}} &
\multicolumn{2}{c}{\textbf{Hand-Crafted}} \\
\cmidrule(lr){3-4}
\cmidrule(lr){5-6}
& &
\textbf{Step Acc. $\uparrow$} &
\textbf{Agent Acc. $\uparrow$} &
\textbf{Step Acc. $\uparrow$} &
\textbf{Agent Acc. $\uparrow$} \\
\midrule
All-at-Once  & Opus-5 & 44.00 & 63.20 & 18.97 & 51.72 \\
Step-by-Step & Opus-5 & 40.80 & 54.40 & 22.40 & 50.00 \\
\midrule
FAMAS~\citep{famas2025}
  & Qwen2.5-72B & 23.81 & 55.56 & \textbf{41.38} & 62.07 \\
AgenTracer-8B~\citep{agentracer2025}
  & Qwen3-8B & 42.86 & 69.62 & 20.68 & \textbf{69.10} \\
CHIEF~\citep{chief2026}
  & DS-V3.2 & 52.00 & \textbf{76.80} & 29.31 & 77.59 \\
CHIEF$^\dagger$~\citep{chief2026}
  & Opus-5 & 44.44 & 61.90 & 27.59 & 63.79 \\
\midrule
\rowcolor{oursgray}
Adaptive Influence Graph & DS-V3.2
  & 50.00 & 66.67 & 22.41 & 60.34 \\
\rowcolor{oursgray}
Adaptive Influence Graph & Sonnet-4
  & 51.59 & 69.05 & 25.86 & 50.00 \\
\rowcolor{oursgray}
Adaptive Influence Graph & GPT-5.6-sol
  & 51.59 & 67.46 & 24.14 & 55.17 \\
\rowcolor{oursgray}
Adaptive Influence Graph & Opus-5
  & \textbf{52.00} & 67.20 & 27.59 & 62.07 \\
\bottomrule
\end{tabularx}

\caption{Who\&When benchmark with ground-truth access at inference time.
Baseline numbers are as reported by their authors. One Algorithm-Generated log
is content-filtered by Opus-5; therefore, we use $n{=}125$ for this model.
$^\dagger$Run using the authors' released code. Our methods are shaded.}
\label{tab:sota_alg_generated_w_gt}
\end{table*}

\section{Representation Ablations Across Backbones}
\label{app:cross_model_ablations}

\Cref{tab:opus_ablations} reports the representation ladder under Opus-5.
\Cref{tab:sonnet_ablations,tab:deepseek_ablations} repeat it under Sonnet-4 and
DS-V3.2. The ordering is the same in all three: the raw transcript is the
weakest interface, structuring it helps, the Influence Graph adds a further
gain, and the AIG is strongest. The size of each step differs---the two weaker
backbones, DS-V3.2 and Sonnet-4, gain far more from structure alone than Opus-5 does, which already
reads the flat transcript comparatively well---but no backbone reverses the
ladder, so the effect of the interface is not an artifact of one model.

\begin{table*}[h]
\centering
\small
\setlength{\aboverulesep}{0pt}
\setlength{\belowrulesep}{0pt}
\renewcommand{\arraystretch}{1.25}
\setlength{\tabcolsep}{3pt}
\begin{minipage}[t]{0.48\textwidth}
\centering
\begin{tabularx}{\linewidth}{l>{\raggedright\arraybackslash}Xcc}
\toprule
\textbf{Model} & \textbf{Representation} & \textbf{Step Acc. $\uparrow$} & \textbf{Agent Acc. $\uparrow$} \\
\midrule
\multirow{4}{*}{Sonnet-4}
  & Raw log        & 20.63 & 61.90 \\
  & Structured log & 48.41 & 60.32 \\
  & IGs            & 49.60 & 66.40 \\
  & \textbf{AIGs}  & \textbf{53.97} & \textbf{67.46} \\
\bottomrule
\end{tabularx}
\caption{Sonnet-4 representation ablations on the Algorithm-Generated
Who\&When partition (no ground-truth access). Our AIGs are in bold.}
\label{tab:sonnet_ablations}
\end{minipage}\hfill
\begin{minipage}[t]{0.48\textwidth}
\centering
\begin{tabularx}{\linewidth}{l>{\raggedright\arraybackslash}Xcc}
\toprule
\textbf{Model} & \textbf{Representation} & \textbf{Step Acc. $\uparrow$} & \textbf{Agent Acc. $\uparrow$} \\
\midrule
\multirow{4}{*}{DS-V3.2}
  & Raw log        & 19.84 & 62.70 \\
  & Structured log & 45.24 & 62.70 \\
  & IGs            & 51.20 & 64.00 \\
  & \textbf{AIGs}  & \textbf{53.17} & \textbf{65.08} \\
\bottomrule
\end{tabularx}
\caption{DeepSeek-V3.2 (DS-V3.2) representation ablations on the
Algorithm-Generated Who\&When partition (no ground-truth access). Our AIGs are
in bold.}
\label{tab:deepseek_ablations}
\end{minipage}
\end{table*}

\section{IG Example}
\label{app:ig_example}

An IG is what the enrichment builder produces: the deterministic node partition
of \Cref{subsec:build} left untouched, with an abstraction written into each
node and inheritance edges added over the chain. Every node carries a short
account of what it did, the input it operated under, and the artifact it
committed, alongside the verbatim log content it abstracts. Edges are added
only where a later node reused part of an earlier node's work in a way tied to
what went wrong. \Cref{fig:ig-json} shows an IG for one trace, where two of the
five edges are non-adjacent and one carries the opening node's reframing of the
task directly into the final verification.

\begin{figure*}[t]
\begin{lstlisting}[
  language=json,
  basicstyle=\ttfamily\scriptsize,
  showstringspaces=false,
  breaklines=true,
  breakautoindent=false,
  columns=fullflexible,
  keepspaces=true,
  linewidth=\textwidth,
  aboveskip=4pt,
  belowskip=4pt
]
{
  "nodes": [
    {
      "id": "A",
      "step_refs": [0],
      "agent": "JSON_Expert",
      "summary": "Received a reframed task: instead of the original DVD-to-Colombia shipping ....",
      "input": "A general task showing 'exitcode: 1 (execution failed), Code output: unknown ....",
      "output": "Restated task framing: debug the 'unknown language json' error following the ....",
      "authorship": "relayed",
      "raw_content": "You are given: (1) a task and advises from your manager with a specific ...."
    },
    {
      "id": "B",
      "step_refs": [1],
      "agent": "Debugging_Expert",
      "summary": "Interpreted 'unknown language json' as a language-code parsing issue. ....",
      "input": "Task from Step 0: analyze and resolve the 'unknown language json' error.",
      "output": "A mock Python script plus a modified version defaulting unsupported languages ....",
      "authorship": "authored",
      "raw_content": "# Analyzing the Error Message and Identifying Potential Causes The error ...."
    },

    ...

    {
      "id": "F",
      "step_refs": [5, 6],
      "agent": "Verification_Expert",
      "summary": "Declared the debugging task resolved based on the successful script execution ....",
      "input": "Successful execution output from Step 4 confirming the script runs without error.",
      "output": "Verification write-up concluding the 'unknown language json' issue is resolved ....",
      "authorship": "authored"
    }
  ],
  "edges": [
    {
      "from": "A",
      "to": "B",
      "inherited": "Step 0 reframed the whole task as a 'debug the unknown language json error' problem, discarding the DVD-shipping ....",
      "effect": "Step 1 inherited that framing and pursued a fabricated language-parsing bug, ensuring the shipping question was ...."
    },
    {
      "from": "B",
      "to": "C",
      "inherited": "Step 1 delivered code blocks whose fence/language tagging was itself malformed.",
      "effect": "Step 2 refused them with 'unknown language json' -- the same symptom under investigation, now reproduced by the ...."
    },
    {
      "from": "C",
      "to": "D",
      "inherited": "Step 2 reported the error literally, as a language-code failure rather than a code-fence tagging issue.",
      "effect": "Step 3 kept treating it as a bug in the script and resubmitted the same fabricated fix."
    },
    {
      "from": "A",
      "to": "F",
      "inherited": "Step 0's reframed objective was to fix the 'unknown language json' error.",
      "effect": "Step 5 declared success on that objective, never returning the DVD-to-Colombia sender/price JSON the original ...."
    },
    {
      "from": "B",
      "to": "D",
      "inherited": "Step 1 fabricated a `parse_language_setting` mock and an interpretation of the error to go with it.",
      "effect": "Step 3 reused both, propagating the misdiagnosis instead of revisiting the framing."
    }
  ]
}
\end{lstlisting}

\caption{\textbf{Example of an IG.} Nodes follow the deterministic partition
and carry an abstraction of the log steps they ground in; edges annotate what
a later node reused from an earlier one and how it went wrong downstream.}
\label{fig:ig-json}
\end{figure*}

\section{AIG Example}
\label{app:aig_example}

An AIG is what the build stage produces: a failed trajectory rewritten as a
compact, query-specific object rather than a transcript to be read end to end.
Consecutive log steps are merged into nodes at the granularity of an
\textit{action} and given a short abstraction of what that action received and
produced, and edges are drawn between nodes---adjacent or several steps
apart---where a later action reused part of an earlier one's work in a way tied
to the failure. The result is a small graph the reader can navigate and
backtrack along, verifying each claim against the steps the node cites.
\Cref{fig:graph-json} shows a finalized AIG for one trace.

\begin{figure*}[t]
\begin{lstlisting}[
  language=json,
  basicstyle=\ttfamily\scriptsize,
  showstringspaces=false,
  breaklines=true,
  breakautoindent=false,
  columns=fullflexible,
  keepspaces=true,
  linewidth=\textwidth,
  aboveskip=4pt,
  belowskip=4pt
]
{
  "nodes": [
    {
      "id": "A",
      "step_refs": [0],
      "agent": "Data_Extraction_Expert",
      "summary": "Received the manager's task framing: count High Energy Physics - Lattice ....",
      "input": "General question about counting hep-lat Arxiv articles from January 2020 with ps ....",
      "output": "Restated plan and constraints; carried forward prior result of 0 with the ....",
      "authorship": "relayed",
      "raw_content": "You are given: (1) a task and advises from your manager with a specific ...."
    },
    {
      "id": "B",
      "step_refs": [1],
      "agent": "Verification_Expert",
      "summary": "Proposed a Python script using arxiv_search with query 'cat:hep-lat AND ....",
      "input": "Plan from Step 0 to extract hep-lat Jan 2020 articles and count those with ps ....",
      "output": "Python code invoking arxiv_search and computing ps_count via substring check ....",
      "authorship": "authored",
      "raw_content": "To begin solving this task, we should follow the given plan step-by-step ...."
    },
    {
      "id": "C",
      "step_refs": [2],
      "agent": "Computer_terminal",
      "summary": "Executed the provided Python script. Execution succeeded (exitcode 0) and printed 0.",
      "input": "Python script from Step 1 querying arxiv_search and counting ps in entry_id.",
      "output": "exitcode 0; stdout: 0.",
      "authorship": "executed",
      "raw_content": "exitcode: 0 (execution succeeded) Code output: 0"
    },
    {
      "id": "D",
      "step_refs": [3, 4],
      "agent": "Verification_Expert",
      "summary": "Interpreted the execution output of 0 as confirmation that no hep-lat January ....",
      "input": "Computer_terminal result of 0 from Step 2.",
      "output": "Final answer 0; TERMINATE signal issued. raw_content (step 3): Based on the ....",
      "authorship": "relayed"
    }
  ],
  "edges": [
    {
      "from": "B",
      "to": "C",
      "inherited": "Step 1 defined ps-detection as substring 'ps' in entry_id, a field that holds an arxiv URL, not format info.",
      "effect": "Step 2 executed this flawed heuristic and produced 0."
    },
    {
      "from": "C",
      "to": "D",
      "inherited": "Step 2 returned 0, a figure produced entirely by the entry_id substring methodology.",
      "effect": "Step 3 accepted that 0 at face value without questioning whether arxiv_search returns any indicator of ps ...."
    },
    {
      "from": "A",
      "to": "D",
      "inherited": "Step 0 propagated the prior '0 / no articles found' framing.",
      "effect": "Step 3 leaned on that same conclusion to justify terminating with 0 rather than probing the ps-detection method."
    }
  ]
}
\end{lstlisting}

\caption{\textbf{Example of an AIG.} The representation the builder produces
and the reader consumes: typed nodes carrying a short abstraction of the log
steps they ground in, and inheritance edges annotating what a later node reused
from an earlier one.}
\label{fig:graph-json}
\end{figure*}

\end{document}